\documentclass[journal]{IEEEtran}
\usepackage{float}
\usepackage{array}
\usepackage{tabularx}
\usepackage{subfigure}
\usepackage{colortbl} 
\usepackage{setspace}
\usepackage{url}
\usepackage{amsmath, bm}
\usepackage{algorithm}  
\usepackage{algorithmic}
\usepackage{colortbl}
\usepackage{stfloats}
\usepackage{graphicx}
\usepackage{hyperref}
\usepackage{booktabs}
\usepackage{multicol}
\usepackage{multirow}
\usepackage{makecell}
\usepackage{amssymb}
\usepackage[table]{xcolor}
\definecolor{gray}{rgb}{.949,.949,.949}
\newcolumntype{Y}{>{\centering\arraybackslash}X}
\hypersetup{hidelinks,colorlinks=false,pdfstartview=Fit,breaklinks=true}

\begin{document}
\title{Applying Large Language Models to Power Systems: Potential Security Threats}

\author{Jiaqi~Ruan,~\IEEEmembership{Member,~IEEE,}
        Gaoqi~Liang,~\IEEEmembership{Member,~IEEE,}
        Huan~Zhao,~\IEEEmembership{Member,~IEEE,}
        Guolong~Liu,~\IEEEmembership{Member,~IEEE,}
        Xianzhuo~Sun,~\IEEEmembership{Member,~IEEE,}
        Jing~Qiu,~\IEEEmembership{Senior Member,~IEEE,}
        Zhao~Xu,~\IEEEmembership{Senior Member,~IEEE,}
        Fushuan~Wen,~\IEEEmembership{Fellow,~IEEE,}
        and~Zhao~Yang~Dong,~\IEEEmembership{Fellow,~IEEE}

\thanks{This work was supported in part by the Research Grants Council of the Hong Kong Special Administrative Region under Grant AoE/P-601/23-N; in part by the General Research Fund of the Hong Kong Special Administrative Region under Grant PolyU15209322; and in part by PolyU under Grant 1-YWCV. \emph{(Corresponding authors: Gaoqi Liang; Zhao Xu.)}}
\thanks{Jiaqi Ruan, Xianzhuo Sun and Zhao Xu are with the Department of Electrical and Electronic Engineering, The Hong Kong Polytechnic University, Hong Kong (e-mail: \href{mailto:jiaqi.ruan@polyu.edu.hk}{jiaqi.ruan@polyu.edu.hk}; \href{mailto:xianzsun@polyu.edu.hk}{xianzsun@polyu.edu.hk}; \href{mailto:eezhaoxu@polyu.edu.hk}{eezhaoxu@polyu.edu.hk}).}
\thanks{Gaoqi Liang is with the School of Mechanical Engineering and Automation, Harbin Institute of Technology, Shenzhen, Shenzhen 518055, China (e-mail: \href{mailto:lianggaoqi@hit.edu.cn}{lianggaoqi@hit.edu.cn}).}
\thanks{Huan Zhao and Guolong Liu are with the School of Science and Engineering, The Chinese University of Hong Kong, Shenzhen, Shenzhen 518172, China (e-mail: \href{mailto:zhaohuan@cuhk.edu.cn}{zhaohuan@cuhk.edu.cn}; \href{mailto:liuguolong@cuhk.edu.cn}{liuguolong@cuhk.edu.cn}).}
\thanks{Jing Qiu is with the School of Electrical and Information Engineering, The University of Sydney, Sydney, NSW 2006, Australia (e-mail: \href{mailto:jeremy.qiu@sydney.edu.au}{jeremy.qiu@sydney.edu.au}).}
\thanks{Fushuan Wen is with the College of Electrical Engineering, Zhejiang University, Hangzhou 310027, China (e-mail: \href{mailto:fushuan.wen@gmail.com}{fushuan.wen@gmail.com}).}
\thanks{Zhao Yang Dong is with the School of Electrical and Electronic Engineering, Nanyang Technological University, Singapore 639798 (e-mail: \href{mailto:zy.dong@ntu.edu.sg}{zy.dong@ntu.edu.sg}).}
}


\maketitle

\begin{abstract}
Applying large language models (LLMs) to modern power systems presents a promising avenue for enhancing decision-making and operational efficiency. However, this action may also incur potential security threats, which have not been fully recognized so far. To this end, this article analyzes potential threats incurred by applying LLMs to power systems, emphasizing the need for urgent research and development of countermeasures. 
\end{abstract}


%

\section{Introduction}\label{sec1}

\IEEEPARstart{I}{n} the dynamically evolving landscape of the power sector, characterized by a growing reliance on renewable energy sources and the integration of diverse grid-connected entities, the complexity and openness of power systems have intensified \cite{wangEventbasedSecureLoad2021}. This evolution presents substantial challenges for power system operators who are tasked with intricate scheduling decisions within an ever-expanding operational scope. As such, the deployment of large language models (LLMs) \cite{porsdammannAUTOGENPersonalizedLarge2023}—sophisticated deep learning frameworks trained on extensive text corpora—has emerged as a transformative solution. These models excel in understanding and generating human-like linguistic expressions, equipping operators with tailored tools for managing complex scenarios more effectively.

The integration of LLMs signifies a significant advancement in addressing the complexities and decision-making challenges of modern power systems. With capabilities in natural language processing, image recognition, and time series analysis \cite{carliniExtractingTrainingData2021}, LLMs can act as a powerful, multifaceted tool for navigating power systems in a complex data milieu. They enhance the extraction of critical information from vast datasets, strengthening the foundations of scheduling and decision-making optimization. Leveraging their robust analytical and logical reasoning capabilities, LLMs facilitate intelligent data retrieval and question-answering, enabling the analysis of various inputs such as historical load data, weather forecasts, and real-time news. This significantly contributes to the formulation of sophisticated optimization strategies. Moreover, LLMs enrich human-computer interactions \cite{wuAIChainsTransparent2022} within power systems, allowing for intuitive presentations of complex data and operational states, thereby supporting operators in making well-informed, high-quality decisions. The application of LLMs in power systems not only enhances dispatch accuracy and efficiency but also highlights their role in improving system adaptability and stability, opening new avenues for research and promising commercial prospects.

However, alongside the growing interest from major power groups in developing tailored LLMs for power systems, this application also introduces significant security concerns \cite{weidingerTaxonomyRisksPosed2022}. While LLMs offer numerous benefits, their deployment within increasingly open power systems can also lead to potential security threats, particularly in data security and decision-making stability. Currently, there is a notable gap in research and investigation into these issues. This article aims to address this gap by providing an in-depth analysis of potential threats posed by LLM applications in power systems, thereby enriching the understanding within both academic and industrial communities and steering the development of more secure LLMs for power system applications.

\section{Potential Threats of Large Language Models in Power Systems}\label{sec2}
\subsection{Large Language Models in Power Systems}
\begin{figure}[t!]
    \centering 
    \includegraphics[width = 1\columnwidth]{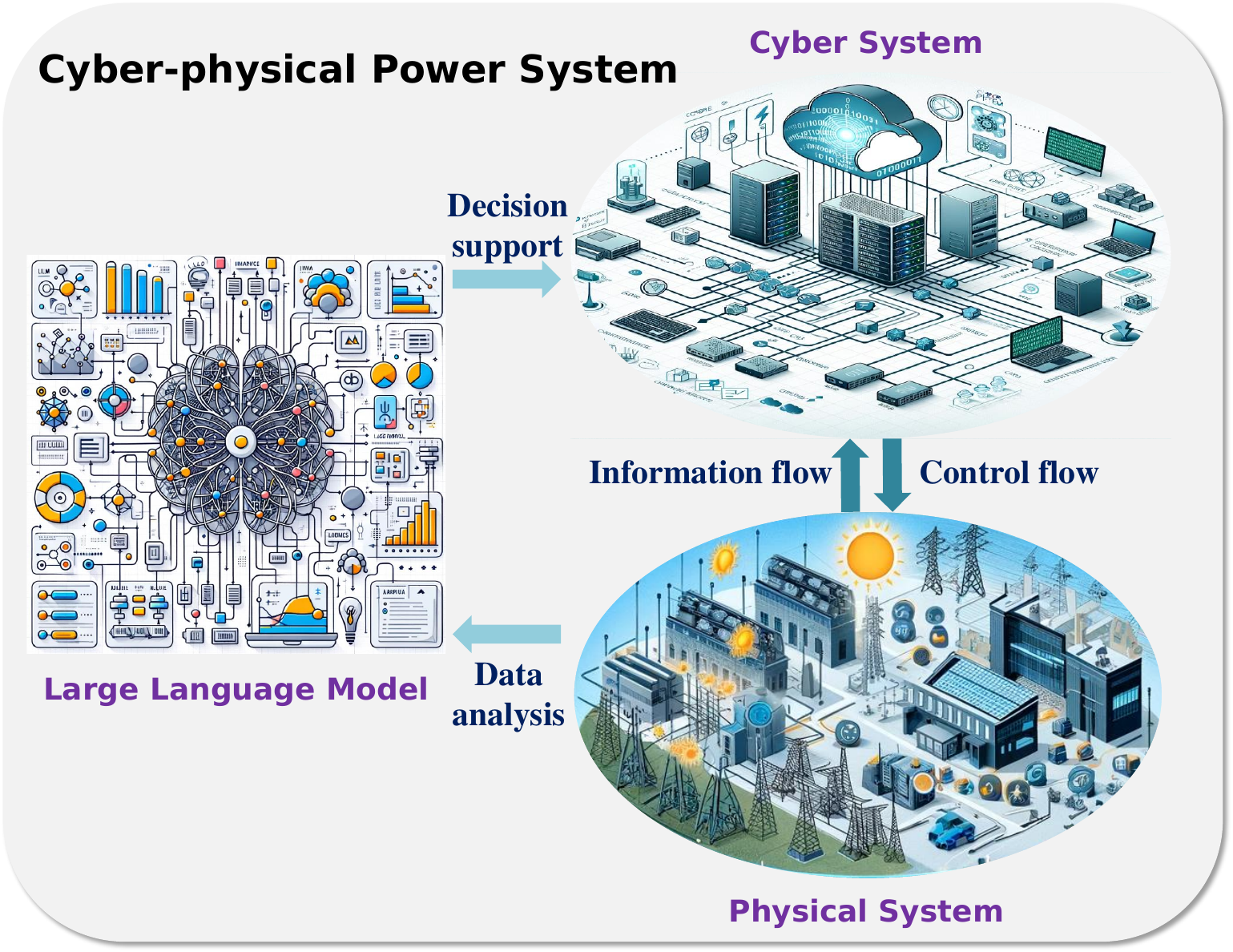}
    \caption{Illustration of applying the large language model to the cyber-physical power system.}
    \label{Illustration} 
\end{figure}

As traditional power systems evolve into cyber-physical power systems (CPPS), the integration of advanced information and communication technology with physical power systems enables real-time perception, rapid response, and intelligent scheduling \cite{wangDynamicDataInjection2019}. In this context, the introduction of LLMs emerges as a crucial advancement towards enhancing the intelligence features of CPPS and optimizing power system operations. As illustrated in Fig. \ref{Illustration}, LLM can play a critical role in analyzing data derived from the physical system, meanwhile aiding in decision-making processes in the cyber system.

Prior to the application of LLMs in power systems, it is required to develop tailored LLMs that synergize advanced artificial intelligence training methodologies with extensive domain-specific knowledge pertaining to power systems. This process initiates with the collection and preparation of a comprehensive dataset, tailored to the industry's requirements, followed by the fine-tuning of a robust base model to understand and generate language contactable to power systems. This necessitates not only technical expertise in machine learning and natural language processing but also a deep understanding of the power system's unique challenges and nuances. Subsequent to the training phase, the focus shifts to rigorous testing and validation to ensure the model's accuracy and practicality in real-world scenarios. With their deep understanding and generation of human language, LLMs provide unique advantages in multimodal data analysis, natural language processing, and intelligent decision support in power systems.

Within the CPPS architecture, LLMs primarily function in the cyber system's information and application layers. In the information layer, they process and analyze substantial data from sensors, smart meters, and other devices. For instance, LLMs can analyze historical and real-time load data, weather information, and user behaviors to forecast electricity demand. In the application layer, LLMs assist in decision-making and optimizing power system operations, generating scheduling strategies and providing valuable insights to power system operators. Furthermore, through natural language processing technology, LLMs can enhance human-machine interaction, contributing significantly to the overall intelligence of CPPS.

Despite these benefits, the integration of LLMs into CPPS may also face potential security threats due to modern power systems' ever-evolving openness and complexity. This is a departure from traditional optimization-based decision-making models in power systems, which are typically characterized by their interpretability, transparency, and specificity. The security threats associated with LLMs include their potential misuse in cyberattacks, susceptibility to data tampering, and implications for system stability. At the same time, the decision-making process of LLMs often lacks transparency, which could impact the stability and security of the power system during critical moments. Therefore, a deeper investigation into these potential threats is essential for the safe operation of CPPS.

\subsection{Potential Threats Incurred by Applying Large Language Models}
The integration of LLMs in power systems may incur various types of security threats, such as data privacy, operational integrity, and system vulnerabilities. These threats could materialize through several mechanisms as follows.
\subsubsection{Privacy invasion through large language models}
The application of LLM in power systems may pose security risks regarding data privacy. Although the advanced data processing and analysis capabilities of LLM significantly improve the operational efficiency of the power system, they may also pose risks for privacy breaches. This is mainly due to LLMs being designed to improve collaboration efficiency across various departments within the power system and often being deployed as resources widely accessible within the system. Such extensive accessibility makes LLMs potential targets for attackers. Once attackers gain access, they can use LLMs' intelligent question-answering system to obtain sensitive information about the power system, such as operational data, control strategies, and even security protocols. This kind of privacy theft not only violates data security but also may enable attackers to launch more complex attacks, such as false data injection attacks (FDIAs). 

Since FDIA was first proposed in 2009, it has been a hot research topic in academia \cite{liangReviewFalseData2017}. However, the implementation of FDIA comes with a strong assumption, wherein attackers have knowledge of the power system's real-time operational conditions, at least partially, to devise effective strategies \cite{ruanSuperResolutionPerceptionAssisted2023}. Traditionally, obtaining such information has been a significant barrier for attackers, making it difficult to fulfill in reality. However, with LLMs in future power systems, this barrier may be significantly lowered. Attackers could use LLMs to obtain detailed operational information and then use this to design FDIA strategies that undermine the power system's stability. Therefore, while LLMs enhance power system intelligence and efficiency, they also introduce privacy breach risks that could be exploited in sophisticated cyberattacks like FDIA. Potential mitigation strategies may include limiting the access of LLMs to sensitive operational data and employing data sanitization techniques. This involves filtering and modifying the operational data in a way that remains useful for legitimate purposes but becomes ineffective for designing FDIA strategies, thereby reducing the security risk of data privacy breaches.

\subsubsection{Deteriorated performance in large language models}
As ultra-large-scale neural networks, LLMs require substantial computational resources and training investment to ensure performance \cite{benderDangersStochasticParrots2021}. Once deployed in power systems, maintaining their performance becomes crucial. However, a shift in the LLM's internal parameters could lead to long-term inappropriate or incorrect decision-making for the power system, which breaches the operational integrity. The threat of such operational integrity-relevant performance degradation may arise from two main aspects.

Firstly, if the data set (including training, validation, and test sets) is maliciously altered during the training or fine-tuning process, LLMs might learn inaccurate or misleading information. Such errors could lead to deviations in the final model parameters, impacting decision-making accuracy and reliability \cite{ruanDeepLearningCybersecurity2023}. Secondly, there is a risk of direct tampering with LLM's internal parameters. If attackers can access and modify these parameters post-deployment \cite{yangAdversarialFalseData}, the LLM's output could significantly deviate from expectations, reducing decision-making effectiveness and potentially leading to serious operational issues.

In both scenarios, deteriorating LLM performance may lead to erroneous decisions in power system operations, threatening system stability and efficiency. Therefore, securing LLM data and model parameters is crucial to prevent performance degradation. This necessitates strict security measures at all training, deployment, and operational stages to uphold LLM's integrity and reliability.

\subsubsection{Threats from semantic divergence}
LLM deployment can coordinate operations across various departments and operators, boosting the overall efficiency of the power system. However, this also means that numerous terminals can communicate with the LLM, generating human-machine interactions and creating many interfaces with a high degree of openness. In such an open environment, some interfaces might be exposed to attackers. As LLMs can interact with a large number of operators through intelligent question-answering, attackers may exploit these exposed interfaces to launch semantic divergence attacks (SDAs), incurring security threats toward systemic vulnerabilities.

SDAs can be carried out in two ways. The first involves altering the LLM's input data ($i.e.$, query semantics) to elicit irrelevant or misleading answers. For example, a query about "real-time load" might be manipulated to produce results for "historical load" instead. The second method involves directly altering LLM's outputs to create divergent answer semantics.

Regardless of the method, SDAs can lead to operators receiving incorrect or misleading information, which could then be used in decision-making for power system operations. This misinformation could significantly affect the reliability and efficiency of the power system. Therefore, monitoring and protecting LLM inputs and outputs is essential in preventing SDAs. It is necessary to implement strict data validation and security protocols within the power system to ensure information accuracy and consistency, preventing attackers from exploiting LLMs as a tool to attack the power system.

\subsubsection{Denial of service for large language models}
Denial of service (DoS) attacks \cite{huseinovicSurveyDenialofServiceAttacks2020} pose a serious threat to LLMs, increasing power systems' security threats in systemic vulnerabilities. These attacks occur when LLMs receive requests exceeding their processing capabilities, rendering the LLM unusable, overloaded, or slow to respond. The DoS attacks can vary, with the most common form being the inundation of the LLM with numerous query requests, depleting computational resources. Beyond ordinary request flooding, attackers might also design particularly complex or lengthy queries, causing the LLM to consume excessive computational resources in processing a single request. The consequences of DoS attacks are severe as they impact the LLM's immediate response capabilities and can paralyze the entire system.

In power systems, this implies that critical decision-support and data analysis functions might be unavailable when needed. For instance, in emergencies, if operators depend on LLMs for rapid response or decision analysis, a DoS attack could cause delayed or erroneous decisions, affecting the power system's stable operation and safety. Moreover, DoS attacks might be used as part of other attack strategies, such as a diversion or to mask more severe attack activities. Therefore, enhancing LLM security, especially against DoS attacks, is crucial for the safe operation of power systems. This may include effective traffic management, monitoring mechanisms, and designing LLMs with the capacity to resist such attacks.

\section{Conclusion and Suggestions}\label{sec3}
While LLMs are expected to significantly enhance the operational efficiency and decision-making capabilities in future power systems, they also introduce new security threats, ranging from data privacy breaches to susceptibility to cyber threats like SDAs and DoS attacks. Addressing these security challenges necessitates a comprehensive, multi-dimensional framework. Fundamental to this is the development of inherently secure LLM architectures, the implementation of sophisticated anomaly detection methodologies, and the establishment of robust LLM frameworks. Compliance with evolving cybersecurity standards and data protection legislation throughout the lifecycle of LLMs is also imperative.

In mitigating these emerging security risks, the adoption of flexible security policies and regulations is crucial, supplemented by human-in-the-loop strategies to fortify LLMs against evolving security threats. Interdisciplinary collaboration and empirical validation through real-world testing are essential in underpinning these strategies. Such endeavors are critical for the progressive adaptation of power systems, facilitating their seamless integration with the advanced capabilities of LLMs while ensuring stringent security and reliability standards.

For stakeholders in the power sector, prioritizing enhanced cybersecurity measures, data protection protocols, and ethical usage guidelines for LLMs is of utmost importance. It is required to foster a culture of security awareness and preparedness, through comprehensive employee training and collaborative efforts with other industries and governmental entities. Regular risk assessments, meticulous monitoring, and periodic system updates are imperative to counter potential security threats associated with LLMs. This holistic framework is essential for applying LLMs to future power systems, aiming to balance the exploitation of LLMs' potential with the mitigation of associated security threats. Continual research, proactive implementation, and the development of secure, transparent LLM systems in alignment with regulatory standards are key to maintaining this equilibrium.

\bibliographystyle{IEEEtran}
\bibliography{Reference}

\end{document}